\documentclass[]{article}
\usepackage[letterpaper]{geometry}
\usepackage{mtsummit2021}
\usepackage{times}
\usepackage{url}
\usepackage{latexsym} 
\usepackage{graphicx}
\usepackage{natbib}
\usepackage{layout}
\usepackage{comment}

\begin{document}

\title{\bf Attentive fine-tuning of Transformers for Translation of low-resourced languages @LoResMT 2021}

\author{\name{\bf Karthik Puranik\(^1\)\hfill Adeep Hande\(^1\)\hfill Ruba Priyadharshini\(^2\)\hfill Thenmozhi Durairaj\(^3\)\\ Anbukkarasi~Sampath\(^4\) \hfill Kingston Pal Thamburaj\(^5\)\hfill Bharathi Raja Chakravarthi\(^6\)} 
        \addr{\(^1\)Department of Computer Science, Indian Institute of Information Technology Tiruchirappalli\\ \(^2\)ULTRA Arts and Science College, Madurai, Tamil Nadu, India\\ \(^3\)Sri Sivasubramaniya Nadar College of Engineering, Tamil Nadu, India\\\(^4\)Kongu Engineering College, Erode, Tamil Nadu, India\\\(^5\)Sultan Idris Education University, Tanjong Malim, Perak, Malaysia\\ \(^6\)Insight SFI Research Centre for Data Analytics, National University of Ireland Galway
        } \\ \addr\{\{karthikp, adeeph\}18c@iiitt.ac.in, rubapriyadharshini.a@gmail.com, theni\_d@ssn.edu.in, anbu.1318@gmail.com, fkingston@gmail.com, bharathi.raja@insight-centre.org\}

}
\maketitle
\pagestyle{empty}

\begin{abstract}
This paper reports the Machine Translation (MT) systems submitted by the IIITT team for the English$\rightarrow$Marathi and English$\Leftrightarrow$Irish language pairs LoResMT 2021 shared task. The task focuses on getting exceptional translations for rather low-resourced languages like Irish and Marathi. We fine-tune IndicTrans, a pretrained multilingual NMT model for English$\rightarrow$Marathi, using external parallel corpus as input for additional training. We have used a pretrained Helsinki-NLP Opus MT English$\Leftrightarrow$Irish model for the latter language pair. Our approaches yield relatively promising results on the BLEU metrics. Under the team name IIITT, our systems ranked 1, 1, and 2 in English$\rightarrow$Marathi, Irish$\rightarrow$English, and English$\rightarrow$Irish respectively. The codes for our systems are published\footnote{\url{https://github.com/karthikpuranik11/LoResMT}}.
\end{abstract}

\section{Introduction}

Today, a large number of text and written materials are present in English. However, with roughly around 6,500 languages in the world\footnote{\url{https://blog.busuu.com/most-spoken-languages-in-the-world/}} \citep{chakravarthi2020leveraging,hande2021domain,sarveswaran2021thamizhimorph}, every native monoglot should not be deprived of this knowledge and information. The manual translation is a tedious job involving much time and human resources, giving rise to Machine Translation (MT). Machine Translation involves the automated translation of text from one language to another by using various algorithms and resources to produce quality translation predictions \citep{articlepathak,krishnamurthy-2015-development,krishnamurthy2019development}. Neural Machine Translation (NMT) brought about a great improvement in the field of MT by overcoming flaws of rule-based and statistical machine translation (SMT) \citep{10.1145/3140107.3140111,achchuthan2015language,parameswari2012development,thenmozhi2018deep,kumar2020tamil}. NMT incorporates the training of neural networks on parallel corpora to predict the likeliness of a sequence of words. sequence-to-sequence neural models (seq2seq) \citep{sutskever2014sequence, kalchbrenner-blunsom-2013-recurrent-continuous} are the widely adopted as the standard approach by both industrial and research communities \citep{Jadhav2020MarathiTE,bojar-etal-2016-findings,cheng-etal-2016-semi}.

Even though NMT performs exceptionally well for all the languages, it requires a tremendous amount of parallel corpus to produce meaningful and successful translations \citep{kumar-etal-2020-unsupervised}. With little research on low resourced languages, finding a quality parallel corpus to train the models can be arduous. The two low-resourced languages worked on in this paper are Marathi (mr) and Irish (ga). With about 120 million Marathi speakers in Maharashtra and other states of India, Marathi is recognized as one of the 22 scheduled languages of India\footnote{\url{https://en.wikipedia.org/wiki/Marathi_language}}. The structural dissimilarity which occurs while translating from English (Subject-Verb-Object) to Marathi (Subject-Object-Verb) or vice versa adds up to issues faced while translation \citep{articlegarje}. The Irish language was recognized as the first official language of Ireland and also by the EU\citep{inproceedingsirish,scannell2007crubadan}. Belonging to the Goidelic language family and the Celtic family\citep{scannell2007crubadan, lynn2015minority}, Irish is also claimed as one of the low resourced languages due to its limited resources by the META-NET report \citep{book, scannell2006machine}.

Our paper represents the work conducted for the LoResMT @ MT Summit 2021\footnote{\url{https://sites.google.com/view/loresmt/}} shared task to build MT systems for the low-resourced Marathi and Irish languages on COVID-19 related parallel corpus. We implement Transformer-based \citep{vaswani2017attention} NMT models to procure BLEU scores\citep{articlebleu} of 24.2, 25.8, and 34.6 in English$\rightarrow$Marathi, Irish$\rightarrow$English, and English$\rightarrow$Irish respectively.
\section{Related works}
Neural Machine Translation has been exhaustively studied over the years \citep{kalchbrenner-blunsom-2013-recurrent-continuous}, with several intuitive approaches involving collective learning to align and translate \citep{bahdanau2016neural}, and a language-independent attention bridge for multilingual translational systems \citep{vazquez-etal-2019-multilingual}. There have been several approaches to NMT, with zero-shot translational systems, between language pairs that have not seen the parallel training data during training \citep{johnson2017googles}. The introduction of artificial tokens has reduced the architectural changes in the decoder \citep{Ha2016TowardMN}. There have been some explorations towards neural machine translation in low resource languages, with the development of a multi-source translational system that targets the English string for any source language \citep{zoph-knight-2016-multi}. 

There has been subsequent research undertaken by researchers for machine translation in low-resource Indian languages. \citeauthor{chakravarthi2021survey} surveyed orthographic information in machine translation, examining the orthography's influence on machine translation and extended it to under-resourced Dravidian languages \citep{chakravarthi2019comparison}. Another approach of leveraging the information contained in rule-based machine translation systems to improve machine translation of low-resourced languages was employed \citep{torregrosa-etal-2019-leveraging}. Several approaches involving the improvement of WordNet for low-resourced languages have been explored \citep{chakravarthi2018improving,chakravarthi2019wordnet}. \citeauthor{chakravarthi2019multilingual} constructed MMDravi, a multilingual multimodal machine translation dataset for low-resourced Dravidian languages, extending it from the Flickr30K dataset, and generating translations for the captions using phonetic transcriptions \citep{u-hegde-etal-2021-uvce}.

There have been relatively fewer approaches experimented with and benchmarked when it comes to translating from Marathi to English and vice versa. \citep{aharoni-etal-2019-massively} tried to build multilingual NMT systems, comprising 103 distinct languages and 204 translational directions simultaneously. \citep{jadhav2020marathi,puranik-etal-2021-iiitt} developed a machine translation system for Marathi to English using transformers on a parallel corpus. Other works on improving machine translation include \citep{10.1007/978-981-16-0401-0_21} proposing an English-Marathi NMT using local attention. The same can be stated to Irish, as it is a poorly resourced language, as the quality of the MT outputs have struggled to achieve the same level as well-supported languages \citep{Dowling2016EnglishTI,Rehm2012TheIL}. In recent years, several researchers tried to overcome the resource barrier by creating artificial parallel data through back-translation \citep{Poncelas2018InvestigatingBI}, exploiting out-of-domain data \citep{imankulova-etal-2019-exploiting}, and leveraging other better-resourced languages as a pivot \citep{dowling-etal-2020-human,wu-wang-2007-pivot}.
\section{Dataset}
We use the dataset provided by the organizers of LoResMT @ MT Summit 2021. The datasets can be found here\footnote{\url{https://github.com/loresmt/loresmt-2021}}. It is a parallel corpus for English and the low resourced language, i.e., Irish and Marathi, mostly containing text related to COVID-19 \citep{ojha-etal-2021-findings}. 

\begin{table}[!h]
\begin{center}
    
\renewcommand{\tabcolsep}{3mm}
\begin{tabular}{|l|r|r|}
\hline
Language pair &  English$\Leftrightarrow$Irish & English$\Leftrightarrow$Marathi\\
\hline

Train &  8,112 & 20,933\\
 
Dev & 502 & 500 \\
Test & 1,000 & 1,000\\
\hline
Total & 9,614 & 22,433\\
\hline
\end{tabular}
\end{center}
\caption{ Number of sentences distribution}\label{tab1}
\end{table}

We have used bible-uedin\footnote{\url{https://opus.nlpl.eu/JRC-Acquis.php}} \citep{Christodoulopoulos2015AMP} an external dataset for Marathi. It is a multilingual parallel corpus dataset containing translations of the Bible in 102 languages\citep{article100} and shows the possibility of using the Bible for research and machine translation. English-Marathi corpus contains 60,495 sentences. CVIT PIB\footnote{\url{http://preon.iiit.ac.in/~jerin/bhasha/}} \citep{Philip_2020} has also been used for the purpose of this research. It contains 1,14,220 parallel corpora for English-Marathi.

\section{Methodology}
\begin{figure}
    \centering
    \includegraphics[width=0.8\linewidth,height=6cm]{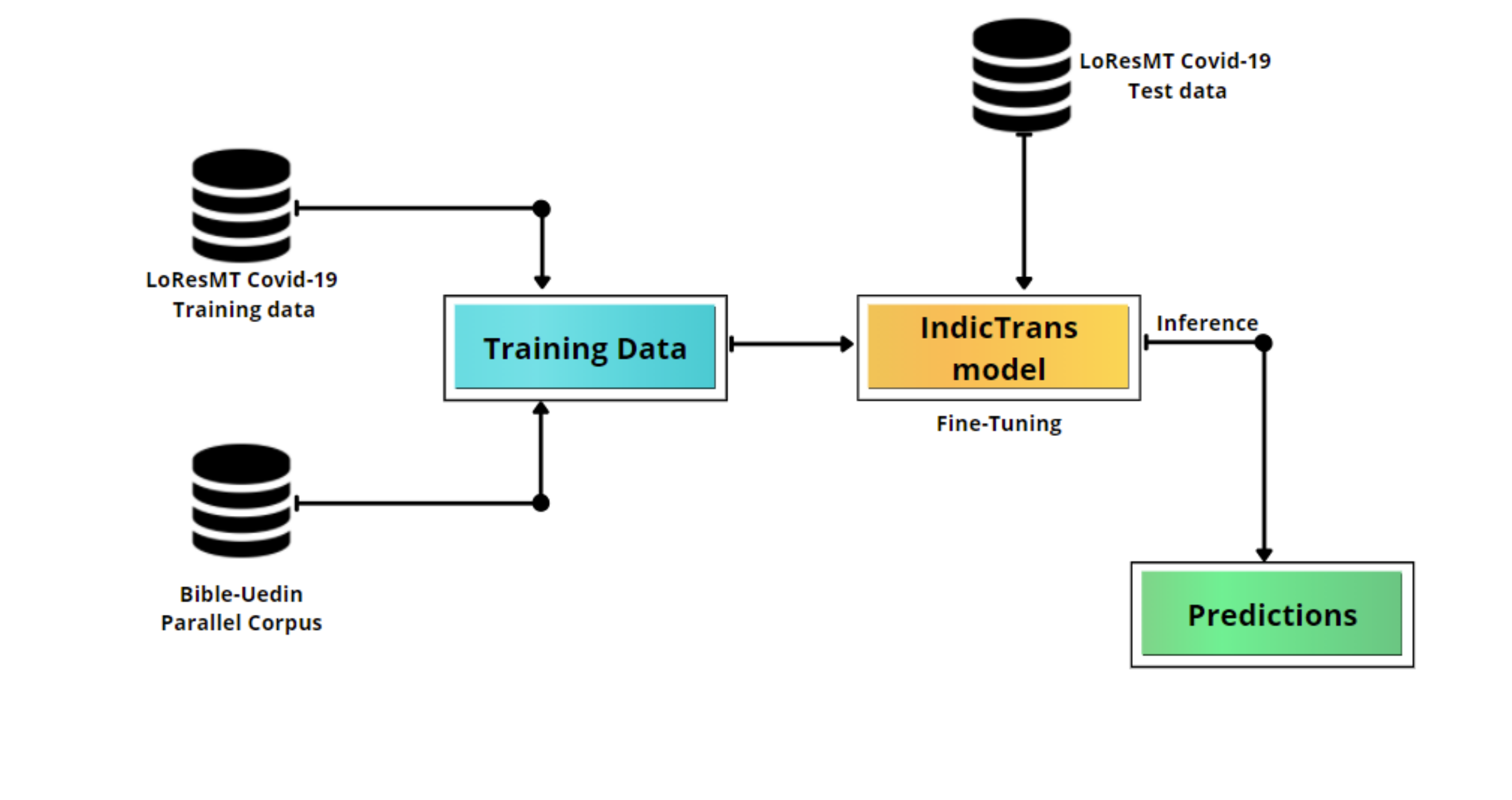}
    \caption{Our approach for the English$\rightarrow$Marathi language pair.}
    \label{fig:approach}
\end{figure}
\subsection{IndicTrans}
Fairseq PyTorch\footnote{\url{https://github.com/pytorch/fairseq}} \citep{ott2019fairseq} is an open-source machine learning library supported as a sequence modeling toolkit. Custom models can be trained for various tasks, including summarization,  language, translation, and other generation tasks. Training on fairseq enables competent batching, mixed-precision training, multi-GPU and multi-machine training. IndicTrans \citep{ramesh2021samanantar}, a Transformer-4x multilingual NMT model by AI4Bharat, is trained on the Samanantar dataset. The architecture of our approach is displayed in Fig.\ref{fig:approach}.
Samanantar\footnote{\url{https://indicnlp.ai4bharat.org/samanantar/}} is the most extensive collection of parallel corpora for Indic languages available for public use. It includes 46.9 million sentence pairs between English and 11 Indian languages. IndicTrans is claimed to successfully outperform the existing best performing models on a wide variety of benchmarks. Even commercial translation systems and existing publicly available systems were surpassed for the majority of the languages. IndicTrans is based on fairseq, and it was fine-tuned on the Marathi training dataset provided by the organizers and the external datasets. The model was fine-tuned with the cross-entropy criterion to compute the loss function, Adam optimizer \citep{zhang2018improved}, dropout of 0.2, fp16 \citep{micikevicius2017mixed}, maximum tokens of 256 for better learning, and a learning rate if 3e-5 in GPU. The process was conducted for a maximum of 3 epochs.

\subsection{Helsinki-NLP Opus-MT}
OPUS-MT \citep{tiedemann-thottingal-2020-opus} supports both bilingual and multilingual models. It is a project that focuses on the development of free resources and tools for machine translation. The current status is a repository of over 1,000 pretrained neural MT models. We fine-tune a transformer-align model that was fine-tuned for the Tatoeba-Challenge\footnote{\url{https://github.com/Helsinki-NLP/Tatoeba-Challenge}}. \textit{\textbf{Helsinki-NLP/opus-mt-en-ga}} model from the HuggingFace Transformers \citep{wolf-etal-2020-transformers} for English$\rightarrow$ Irish and \textit{\textbf{Helsinki-NLP/opus-mt-ga-en}} for Irish$\rightarrow$ English were used.

\begin{table}[htpb]
\begin{center}
    
\renewcommand{\tabcolsep}{3mm}
\begin{tabular}{|l|r|r|}
\hline
Language pair &  Method & BLEU\\
\hline

English$\rightarrow$Marathi & IndicTrans baseline & 14.0 \\
English$\rightarrow$Marathi & IndicTrans TRA & 17.8\\
English$\rightarrow$Marathi & IndicTrans CVIT-PIB & 23.4\\
English$\rightarrow$Marathi & IndicTrans bible-uedin & 27.7\\
\hline
English$\rightarrow$Irish & Opus MT & 30.4\\
English$\rightarrow$Irish & M2M100 & 25.6\\
\hline
Irish$\rightarrow$English & Opus MT & 37.2\\
Irish$\rightarrow$English & M2M100 & 30.4\\

\hline
\end{tabular}
\end{center}
\caption{BLEU scores obtained for the various models for the development set}\label{tab2}
\end{table}
\begin{figure}
    \centering
    \includegraphics[width=0.8\linewidth, height=8cm]{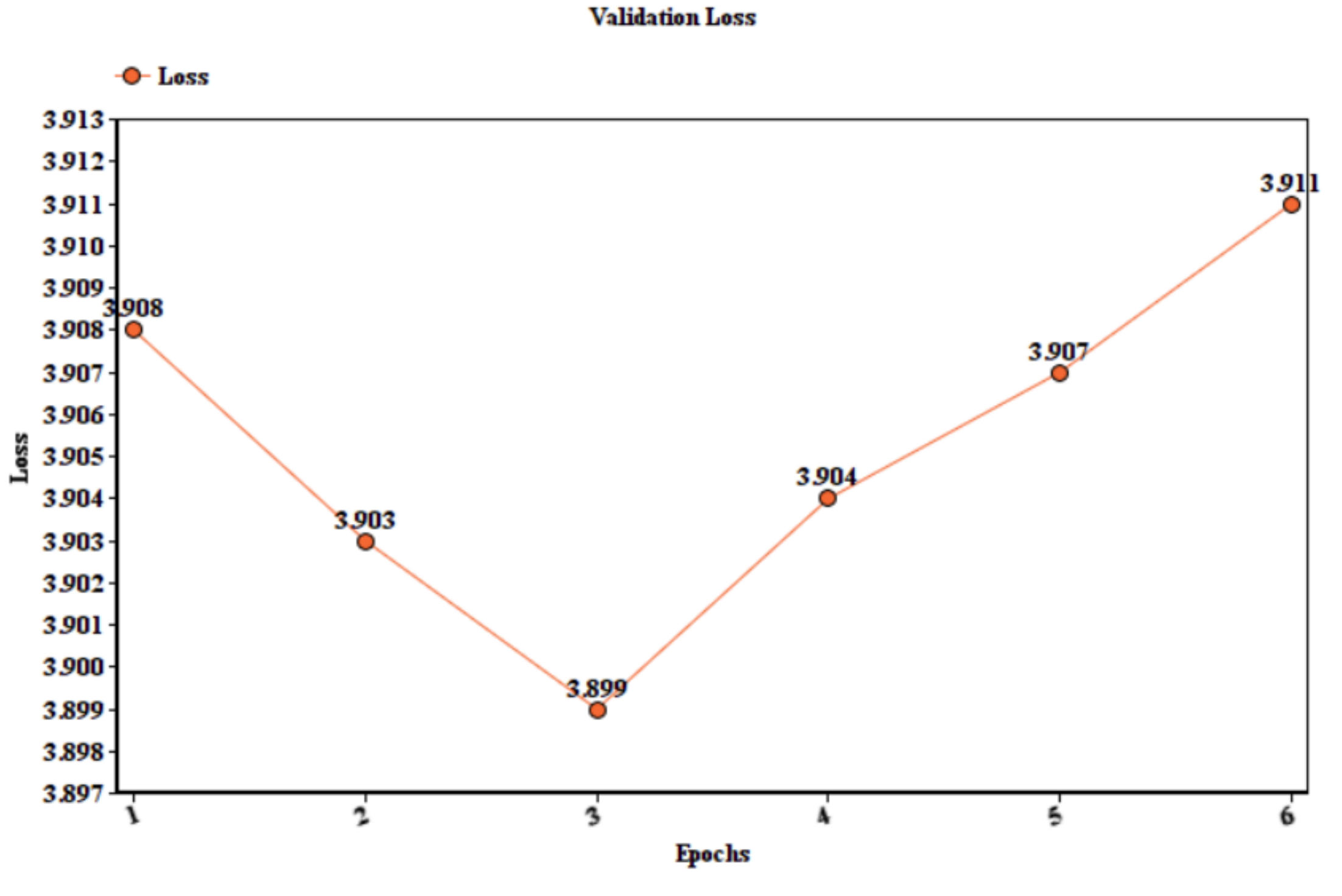}
    \caption{The graph depicting the increase in val loss
    after the third epoch}

    \label{graph}
\end{figure} 
\section{Results and Analysis}
For Marathi, it is distinctly visible that our system model, i.e., IndicTrans fine-tuned on the training data provided by the organizers or TRA and the bible-uedin dataset, gave the best BLEU scores. It was surprising how the model fine-tuned on a parallel corpus of 60,495 sentences of bible-uedin surpassed the model fine-tuned on 1,14,220 sentences from the CVIT PIB dataset. The possible explanation is the higher correlation between the sentences of the bible-uedin dataset with the test dataset than the CVIT PIB dataset. Another reason could be the presence of excessive noise in the CVIT PIB dataset. The other reason for this could be noise and a lower quality of translations in the CVIT PIB dataset compared to bible-uedin.

To infer the same, 1000 random pairs of sentences were picked from the datasets, and the average  LaBSE or language-agnostic BERT Sentence Embedding \citep{feng2020languageagnostic} scores were found out. LaBSE gives a score between 0 and 1, depending on the quality of the translation. It was seen that the average was 0.768 for the bible-uedin dataset, while it was 0.58 for the CVIT PIB dataset. This might have been one of the reasons for the better BLEU scores. The model also showed constant overfitting after the second and third epoch, as the BLEU scores reduced considerably as they reached the 6th epoch. The BLEU scores decreased by a difference of 6.
The validation loss starts increasing after the third epoch, thus, showing the overfitting occurring in training. So, the model was fine-tuned for three epochs while maintaining a low learning rate of around 3e-5 to get a BLEU score of 24.2. 

\begin{table}[!h]
\begin{center}
    
\renewcommand{\tabcolsep}{3mm}
\begin{tabular}{|l|r|r|r|r|}
\hline
Language pair &  BLEU & CHRF & TER & Rank\\
\hline

English$\rightarrow$Marathi & 24.2 & 0.59 & 0.597 & 1\\
Irish$\rightarrow$English & 34.6 & 0.61 & 0.711 & 1\\
English$\rightarrow$Irish & 25.8 & 0.53 & 0.629 & 2\\

\hline
\end{tabular}
\end{center}
\caption{Result and ranks obtained for the test dataset\citep{popovic-2015-chrf, articleter}}\label{tab3}
\end{table}

Training a model to predict for a low-resourced language was highly challenging due to the absence of prominent pretrained models \citep{kalyan-etal-2021-iiitt,yasaswini-etal-2021-iiitt,9418446}. However, as an experiment, two models from HuggingFace Transformers\footnote{\url{https://huggingface.co/transformers/}}, M2M100 \citep{fan2020englishcentric} and Opus-MT from Helsinki NLP \citep{tiedemann2020tatoeba} were compared. For the dev data, Opus MT produced a BLEU score of 30.4 while M2M100 gave 25.62 for translations from English to Irish and 37.2 and 30.37 respectively for Irish to English translations. Probably, the individual models pretrained on numerous datasets gave Opus MT an edge over M2M100. This led us to submit the Opus MT model for the LoResMT Shared task 2021. The model gave exceptional BLEU scores of 25.8 for English to Irish, which ranked second in the shared task, while 34.6 for Irish to English stood first.

\section{Conclusion}
It is arduous and unyielding to get accurate translations for low-resourced languages due to limited datasets and pretrained models. However, our paper puts forward a few methods to better the already existing accuracies. Ranked 1, 1, and 2 in English$\rightarrow$Marathi, Irish$\rightarrow$English, and English$\rightarrow$Irish respectively in the LoResMT 2021 shared task, IndicTrans fine-tuned on the bible-uedin, and the dataset provided by the organizers manages to surpass the other models due to its high correlation with the test set and minimal noise for the Marathi language. The Irish language task was dominated by the Opus MT model by Helsinki-NLP, outperforming other Transformer models, M2M100.

\small

\bibliographystyle{apalike}
\bibliography{mtsummit2021,mtsummit2021_1}

\end{document}